\documentclass[entropy,article,accept,pdftex,moreauthors]{Definitions/mdpi}

\firstpage{1} 
\pubvolume{1}
\issuenum{1}
\articlenumber{0}
\pubyear{2024}
\copyrightyear{2024}
\datepublished{ } 
\hreflink{https://doi.org/} 

\makeatletter
\let\c@lofdepth\relax
\let\c@lotdepth\relax
\makeatother
\usepackage{subfigure}
\usepackage[normalem]{ulem}
\usepackage{makecell}
\newcolumntype{"}{@{\hskip\tabcolsep\vrule width 1pt\hskip\tabcolsep}}

\usepackage{adjustbox}
\usepackage{placeins} 

\newlength{\figwidth}
\Title{Punctuation patterns in \textit{Finnegans Wake} by James Joyce are largely translation-invariant}

\TitleCitation{Punctuation patterns in \textit{Finnegans Wake} by James Joyce are largely translation-invariant}

\Author{Krzysztof Bartnicki $^{1}$, Stanisław Drożdż $^{2,3,}$*\orcidC{}, Jarosław Kwapień $^{2}$\orcidD{} and Tomasz Stanisz $^{2}$\orcidE{}}

\AuthorNames{Krzysztof Bartnicki, Stanisław Drożdż, Jarosław Kwapień and Tomasz Stanisz}

\AuthorCitation{Bartnicki, K.; Drożdż, S.; Kwapień, J.; Stanisz, T.}

\address{%
$^{1}$ \quad Independent researcher, \mbox{Poland;} gimcbart@gmail.com (K.B.)\\
$^{2}$ \quad Complex Systems Theory Department, Institute of Nuclear Physics, Polish Academy of Sciences, \mbox{31-342 Krak\'ow, Poland;} jaroslaw.kwapien@ifj.edu.pl (J.K.); tomasz.stanisz@ifj.edu.pl (T.S.)\\
$^{3}$ \quad Faculty of Computer Science and Telecommunications, Cracow University of Technology, \mbox{31-155 Krak\'ow, Poland;}
}

\corres{Correspondence: stanislaw.drozdz@ifj.edu.pl (S.D.)}

\abstract{The complexity characteristics of texts written in natural languages are significantly related to the rules of punctuation. In particular, the distances between punctuation marks measured by the number of words quite universally follow the family of Weibull distributions known from survival analyses. However, the values of two parameters marking specific forms of these distributions distinguish specific languages. This is such a strong constraint that the punctuation distributions of texts translated from the original language into another adopt quantitative characteristics of the target language. All these changes take place within Weibull distributions such that the corresponding hazard functions are always increasing. Recent previous research shows that James Joyce's famous \textit{Finnegans Wake} is subject to such extreme distribution from the Weibull family that the corresponding hazard function is clearly decreasing. At the same time, the distances of sentence ending punctuation marks, determining the variability of sentence length, have an almost perfect multifractal organization, so far to such an extent found nowhere else in the literature. In the present contribution based on several available translations (Dutch, French, German, Polish, Russian) of \textit{Finnegans Wake}, it is shown that the punctuation characteristics of this work remain largely translation invariant, contrary to the common cases. These observations may constitute further evidence that \textit{Finnegans Wake} is a translinguistic work in this respect as well, in line with Joyce's original intention.}


\begin{document}

\section{Introduction}

A complexity science perspective offers valuable insights into the study of natural language, as multiple traits identify it as a complex system~\cite{HebertDufresne2024,KwapienJ-2012a}. Methods widely employed in analyzing complex systems — rooted in concepts from information theory~\cite{Debowski2020,Takahira2016,Montemurro2011}, time series analysis~\cite{AlvarezLacalle2006,Liu2023,Sanchez2023,Pawlowski1997,KosmidisK-2006a}, network science~\cite{CanchoRF-2001a,AmancioDR-2008a,LiuH-2011a,CongJ-2014a,WachsLopezGA-2016a,KuligA-2017a,AkimushkinC-2017a,StaniszT-2019a,RaduchaT-2018a}, and power-law probability distributions~\cite{NarananS-1998a,Newman2005,AusloosM-2010a,Piantadosi2014} — have been utilized to explore the quantitative characteristics of natural language. Understanding the mechanisms underlying these characteristics can significantly enhance natural language processing (NLP) and generation techniques, a particularly pertinent goal in the era of large language models (LLMs)~\cite{ShanahanM-2023a,ZhaoWX-2023a}, which are foundational to generative AI systems such as OpenAI's ChatGPT, Microsoft's Bing Chat, and Google's Bard.

One language property recently analyzed using statistical approaches is punctuation usage in written texts. Research demonstrates that the distribution of word counts between consecutive punctuation marks in literary texts generally follows a discrete Weibull distribution~\cite{StaniszT-2023a,StaniszT-2024a,DecJ-2024a}. Remarkably, the two parameters of this Weibull distribution are largely language-specific, reflecting the distinctive punctuation characteristics of different languages. Moreover, translations of texts into another language in general retain the Weibull distribution, but with parameter values corresponding to the target language~\cite{StaniszT-2023a,StaniszT-2024a}. Sentence-ending punctuation, such as periods, deviates from this rigidity, however. The intervals thus defined, representing sentence lengths, exhibit greater variability and are not as strictly bound by the Weibull distribution, allowing for a more diverse range of patterns.

In this context \textit{Finnegans Wake} by James Joyce is a challenging object for quantitative scientific analyses as it constitutes one of the most complex and enigmatic works of literature in the English language, known for its experimental style, inventive language that convolutes English with dozens of other languages, and it thus induces intricate, layered meanings. Many words have here multiple meanings, inviting readers to find several interpretations for each word, phrase and sentence. \textit{Finnegans Wake} is structured to reflect the logic of a dream rather than a conventional narrative. The book begins and ends mid-sentence, creating a circular structure where the last line connects back to the first, symbolizing an endless cycle. There is no clear plot or sequence; instead, the narrative spirals, looping back on itself with characters, symbols, and themes that morph and shift. It is supposed to explore how language mirrors human consciousness and the complexity of memory, experience, and culture. Already from this perspective, it is natural to anticipate the appearance of various types of long-range correlations in the text of this book, which do not exist in traditional written texts. 

Indeed, \textit{Finnegans Wake}'s punctuation patterns, which reflect mutual arrangement in the organization of phrases and sentences, contain correlations incomparably more amazing than in any other work that has ever been subjected to this type of analysis. The most spectacular dimension of \textit{Finnegans Wake} manifests itself in the full glory multifractality in the patterns of sentence lengths variability, largely paralleling the model mathematical cascades. The second unusual property of this work is that the longer the sequence of words uninterrupted by punctuation, the less likely it is that a punctuation mark will appear after the next word, which is a unique property among books. In the formal language of probabilistic survival analysis, this means that the hazard function expressing the need for a punctuation mark to appear if it has not been present in the sequence of words is decreasing. It is as if for a living organism the probability of surviving subsequent years increases with the age of that organism even at its age of maturity.

\FloatBarrier

\section{Materials and Methods}

\subsection{\textit{Finnegans Wake}}
James Joyce's \textit{Finnegans Wake} is a unique and challenging work of modernist literature, known for its complex structure, innovative use of language, and unconventional narrative techniques. The opening sentence is a continuation of the final sentence, creating a loop that reflects the cyclical themes in the book. Influenced by Giambattista Vico's theory of historical cycles, the book explores patterns of rise, fall, and renewal across human history. The narrative mimics the structure of a dream, delving into subconscious associations and symbolic imagery. Central to the story is the Earwicker family, representing archetypes of father, mother, and children. The work draws on mythological, biblical, and literary sources, embedding archetypes like the trickster, the hero, and the everyman. The text eschews traditional plot for a fragmented, cyclical approach, mirroring the logic of dreams~\cite{}. 

The book is divided into four parts. Part I (8 chapters) introduces the central characters and themes. It sets the stage for the dreamlike narrative, blending personal and universal histories. Part II (4 chapters) focuses on the Earwicker family, including explorations of their interpersonal dynamics and symbolic representations of societal archetypes. Part III (4 chapters) contrasts rationality and structure versus creativity and chaos of the two brothers. Finally, Part IV (1 chapter) acts as a coda, bringing together the cyclical motifs of renewal and closure as ALP's (Anna Livia Plurabelle) monologue leads back to the beginning.

\textit{Finnegans Wake} is often described as one of the most difficult works of literature due to its linguistic density and abstract narrative. It has been interpreted as a "universal dream," capturing the collective unconscious and human experience across time and space. The text is rich with references to world literature, history, philosophy, and folklore. Sentences often break grammatical rules, defying conventional readability but inviting interpretive exploration. For all these reasons \textit{Finnegans Wake} is often described as one of the most difficult works of literature due to its linguistic density and abstract narrative. 

Translating James Joyce’s \textit{Finnegans Wake} into other languages presents thus an array of unique challenges due to its experimental nature and narrative full of obscurities, ambiguities, and polyglot infusions. The text often prioritizes sound and rhythm over conventional grammar, creating a musicality that is integral to its meaning. Translating this musical quality into another language, while maintaining the sense of the text, is a delicate balancing act. Translators face the challenge of balancing fidelity to Joyce’s complexity with readability for their audience.  While \textit{Finnegans Wake} is intentionally opaque, readers in the target language may lack the cultural or linguistic tools that English-speaking readers can rely on. Joyce’s narrative incorporates multiple voices, tones, and registers, often blending them seamlessly. Translating this heteroglossia is difficult, especially in languages with less flexibility in tonal or dialectal variation. These are the reasons why available complete translations of this book into other languages are not very numerous but exist for several European languages. In addition to the original, five such widely recognized translations are used in the quantitative analysis presented here. They include translation into Dutch~\cite{FW_Dutch}, French~\cite{FW_French}, German~\cite{FW_German}, Polish~\cite{FW_Polish} and Russian~\cite{FW_Russian}.

\subsection{Discrete Weibull distribution}

Punctuation can be understood as a mechanism for interrupting the continuous flow of words, enhancing the clarity of the message and providing necessary pauses for the reader. These functions align well with the framework of survival analysis~\cite{MillerR-1997a}. Prior studies have established that distances between punctuation marks follow distributions that can be modeled using the discrete Weibull distribution~\cite{StaniszT-2023a,StaniszT-2024a}. This distribution is defined by the following probability mass function~\cite{NakagawaT-1975a}:
\begin{equation}
f(k) = (1-p)^{k^{\beta}} - (1-p)^{(k+1)^{\beta}}, \quad p \in (0,1), \quad \beta>0
\label{eq::discrete.weibull.pmf}
\end{equation}
and its cumulative distribution function:
\begin{equation}
\mathcal{F}(k)=1 - \left( 1-p \right)^{k^\beta}.
\label{eq::Weibull.CDF}
\end{equation}

The cumulative distribution function represents the probability that the random variable exceeds the value $k$. The discrete Weibull distribution generalizes the geometric distribution, which is recovered when $\beta = 1$, resulting in a constant probability $\mathcal{F}(k)$ over time. When $\beta > 1$, the probability increases with time, while for $\beta < 1$, it decreases. This distribution has broad applications in fields such as survival analysis, weather prediction, and textual data analysis~\cite{JohnsonNL-1994a,MillerR-1997a,AltmannEG-2009a}. In the context of punctuation, $f(k)$ denotes the probability that a punctuation mark will appear precisely after $k$ words. 

A complementary approach to the above generalization can be readily and transparently formulated in terms of the hazard function $\lambda(k)$ expressing the conditional probability that the $k$th trial will result in a success provided that no success has occurred in the preceding $k-1$ trials:
\begin{equation}
\lambda(k) = \frac{f(k)}{1-\mathcal{F}(k-1)}.
\label{eq::hazard_function}
\end{equation}
In the case of the discrete Weibull distribution it becomes~\cite{PadgettWJ-1985a}:
\begin{equation}\label{eq::hazard_Weibull}
\lambda(k) = \; 1-\left(1-p\right)^{k^{\beta} - (k-1)^{\beta}}.
\end{equation}
For data that exactly follows a Weibull distribution $\beta > 1$ corresponds to $\lambda(k)$ which is an increasing function of $k$. In other words the probability of success increases with the number of preceding unsuccessful trials. The opposite applies to $\beta < 1$. In the \textit{memoryless} case of $\beta = 1$, the hazard function is constant. Since $p=\lambda(1)$ the parameter $p$ reflects the probability of putting a punctuation mark right after the first word following the last punctuation mark. 

\subsection{Multifractal detrended fluctuation analysis (MFDFA)}

Self-similarity, or the absence of a characteristic scale, is a defining feature of natural complex systems. Empirically, this property often manifests as a non-trivial temporal organization in measurement outcomes, represented as a time series. Specifically, complexity is frequently linked to a cascade-like hierarchy of data points exhibiting multiscaling, underscoring the importance of practical methods for identifying such structures in complex systems research~\cite{JimenezJ-2000a}. Drawing on current research, multifractal detrended fluctuation analysis (MFDFA) has proven to be a highly reliable method for analyzing multiscaling structures~\cite{KantelhardtJ-2002a,OswiecimkaP-2006a}. This technique builds upon the widely adopted detrended fluctuation analysis (DFA)~\cite{PengCK-1994a}, offering a multiscale approach to studying hierarchical and multiscaling patterns. Below, we provide a concise outline of the key steps involved in the MFDFA algorithm.

Consider a time series $U=\{u_i\}_{i=1}^T$, consisting of $T$ consecutive measurements of an observable $u$. This series is divided into $M_s$ non-overlapping windows, each of length $s$, starting from both ends of $U$, resulting in a total of $2 M_s$ windows. To address potential non-stationarity in the signal, a detrending procedure is applied within each window to an integrated signal (referred to as the profile) $X=\{x_i\}_{i=1}^s$, with elements defined as:
\begin{equation}
x_i = \sum_{j=1}^i u_j.
\end{equation}
The detrending process involves fitting a polynomial $P^{(m)}$ of order $m$ (with $m=2$ used throughout this study) to the data $X$ within each window $\nu=0,\ldots,2 M_s-1$. The variance of the detrended signal obtained after subtracting the fitted polynomial is then computed:
\begin{equation}
f^2(\nu,s) = {1 \over s} \sum_{i=1}^s (x_i - P^{(m)}(i))^2.
\label{eq::variance}
\end{equation}
In the next step, a family of fluctuation functions of order $q$ is defined using the average variance computed across all windows:
\begin{equation}
F_q(s) = \left\{ {1 \over 2M_s} \sum_{\nu=0}^{2M_s-1} \left[ f^2(\nu,s) \right]^{q/2} \right\}^{1/q}.
\label{eq::fluctuation.functions}
\end{equation}
Here, $q$ is a real number. The fluctuation functions $F_q(s)$ are computed for various values of the scale $s$ and the index $q$. Typically, the minimum $s$ is selected to exceed the length of the longest sequence of constant values in $U$, while the maximum ss is set to $T/5$. Unlike $s$, there is no standard range for $q$. Since $q$ is associated with the moments of the signal, extreme values should be avoided for time series with heavy tails to ensure meaningful results. If the fluctuation functions depend on $s$ as power laws:
\begin{equation}
F_q(s) \sim s^{h(q)}
\label{eq::scaling}
\end{equation}
for a number of different choices of $q$, it indicates that the time series under study is either monofractal (when $h(q)$ is constant in $q$) or multifractal otherwise. The function $h(q)$ is called the generalized Hurst exponent, because for $q=2$, $h(q)=H$, where $H$ is the standard Hurst exponent~\cite{HurstHE-1951a,HeneghanC-2000a}. From a visual perspective, fractal $F_q(s)$ organizations result in straight lines on double logarithmic plots.

A practical representation of the multifractal characteristics of data is the singularity spectrum $f(\alpha)$, which can be derived from $h(q)$ by the Legendre transform:
\begin{eqnarray}
\nonumber
\alpha = h(q) + qh'(q),\\
f(\alpha) = q \left[\alpha-h(q)\right] + 1,
\label{eq::singularity.spectrum}
\end{eqnarray}
Here, $\alpha$ represents the measure of data-point singularity, equivalent to the H\"older exponent. Geometrically, $f(\alpha)$ can be interpreted as the fractal dimension of the subset of the data characterized by a specific H\"older exponent ${\alpha}$~\cite{HalseyTC-1986a}. For a monofractal time series, the pair $(\alpha,f(\alpha))$ reduces to a single point. In contrast, for a multifractal time series, it typically forms a downward-pointing parabola. The broader the singularity spectrum $f(\alpha)$, the greater the richness of multifractality in the time series, which serves as an indicator of its complexity content. In some cases, the $f(\alpha)$ parabola may appear distorted or asymmetric, suggesting that data points of varying amplitudes exhibit different scaling behaviors~\cite{OhashiK-2000a,CaoG-2013a,DrozdzS-2015a,GomezGomezJ-2021a,KwapienJ-2022a}.

\section{Results and Discussion}

\subsection{Inter-punctuation intervals (IPI)}

\begin{figure}[p]
\newlength{\figDistrWidth}
\setlength{\figDistrWidth}{0.45\textwidth}
\newlength{\figDistrSpace}
\setlength{\figDistrSpace}{2em}
\centering

\includegraphics[width=\figDistrWidth]{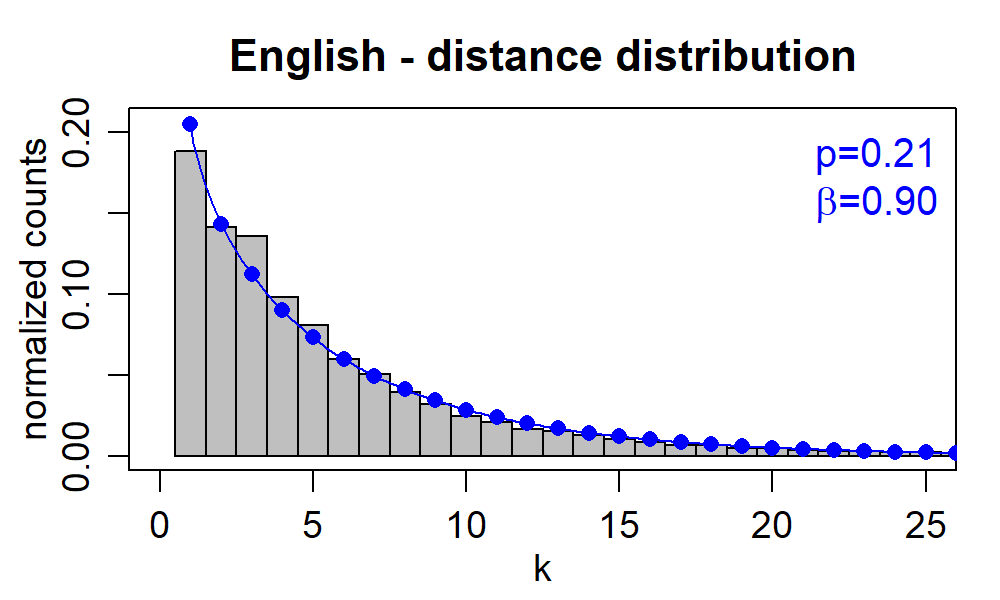}
\hspace{\figDistrSpace}
\includegraphics[width=\figDistrWidth]{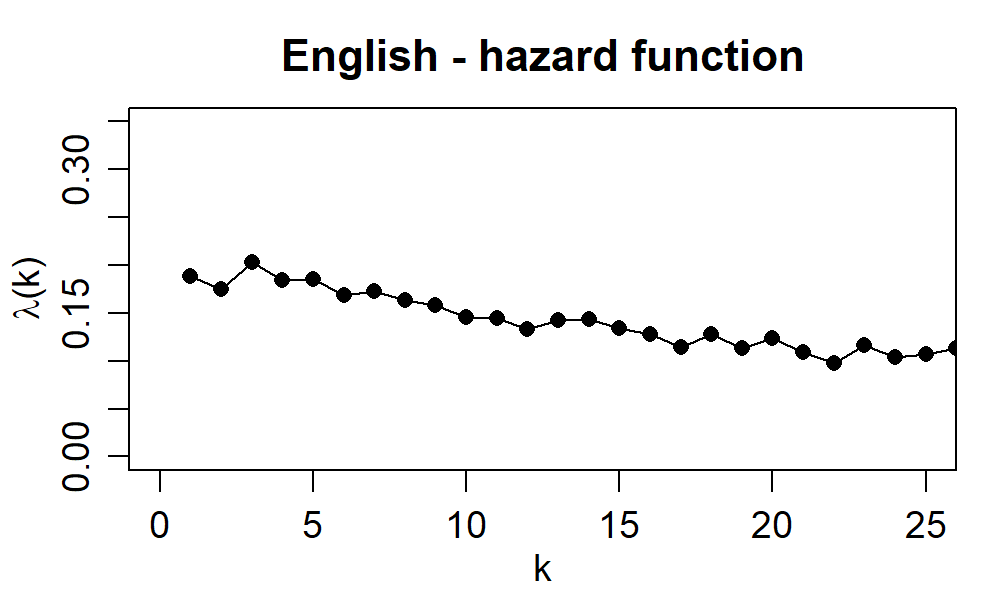}

\includegraphics[width=\figDistrWidth]{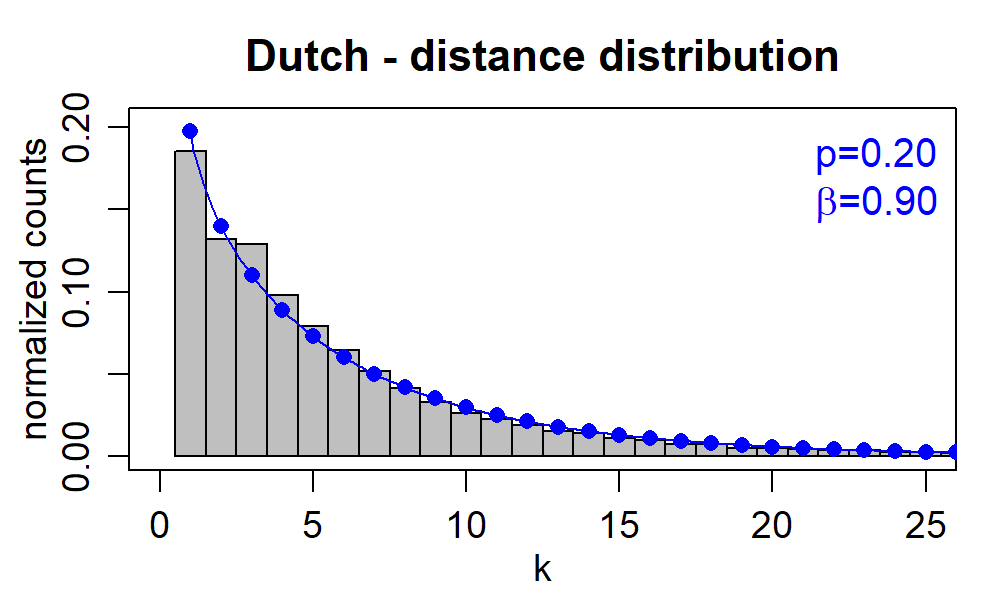}
\hspace{\figDistrSpace}
\includegraphics[width=\figDistrWidth]{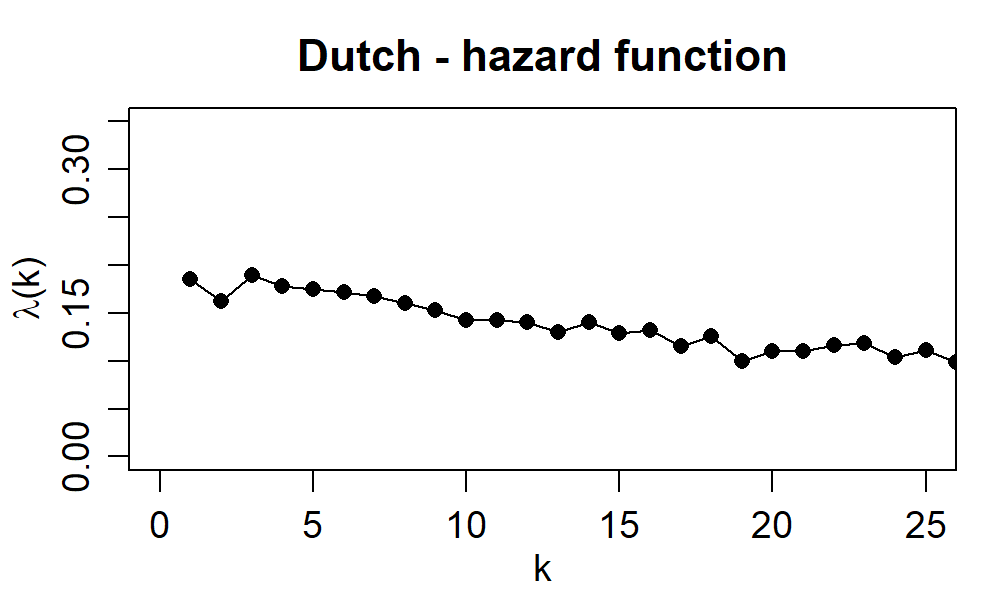}

\includegraphics[width=\figDistrWidth]{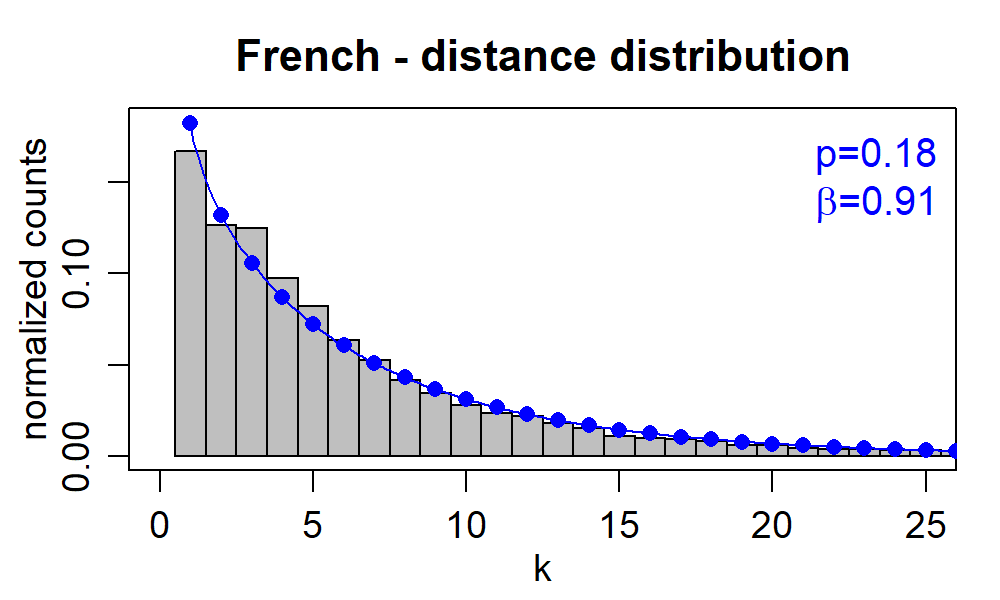}
\hspace{\figDistrSpace}
\includegraphics[width=\figDistrWidth]{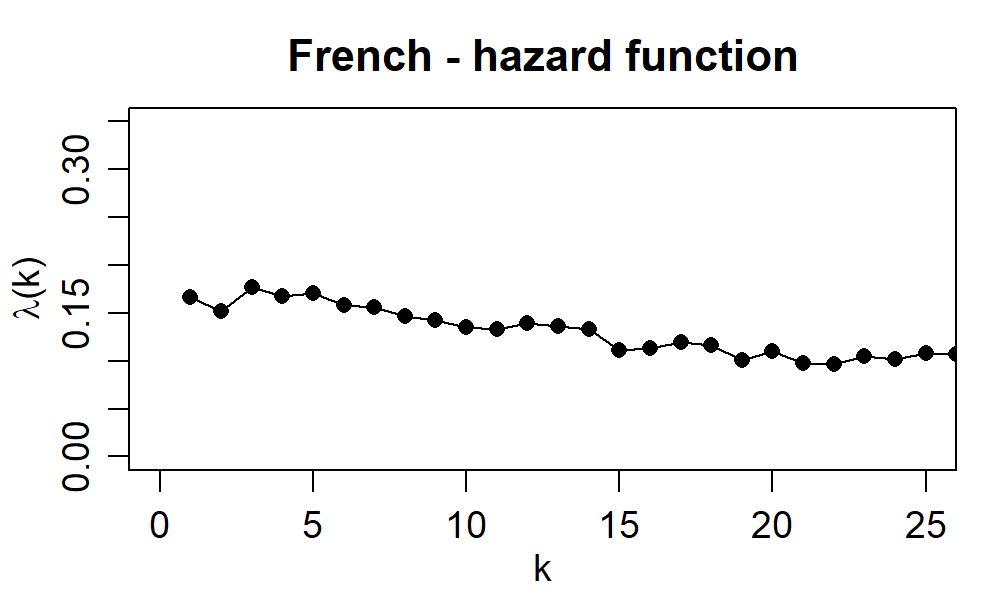}

\includegraphics[width=\figDistrWidth]{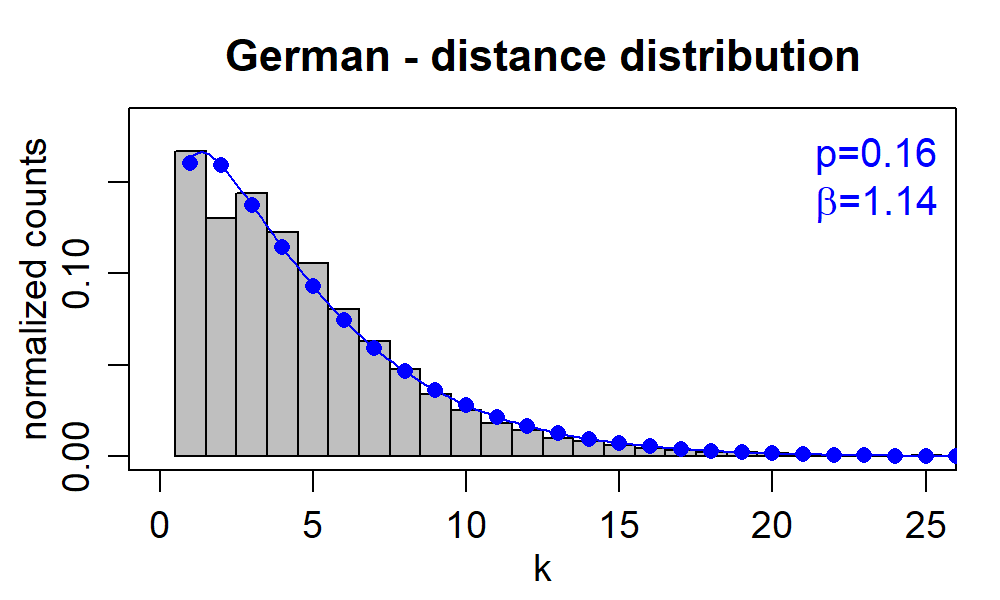}
\hspace{\figDistrSpace}
\includegraphics[width=\figDistrWidth]{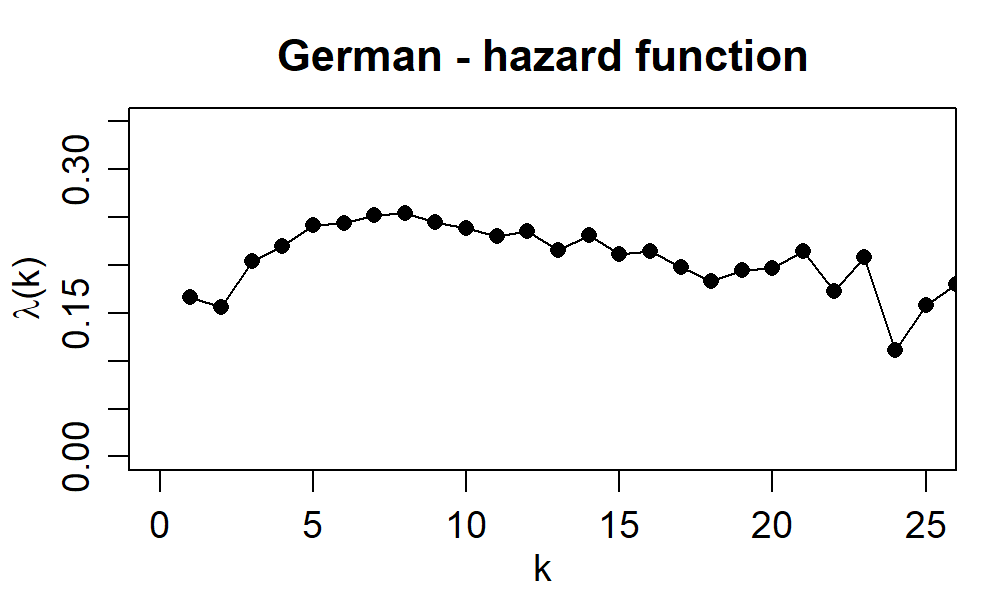}

\includegraphics[width=\figDistrWidth]{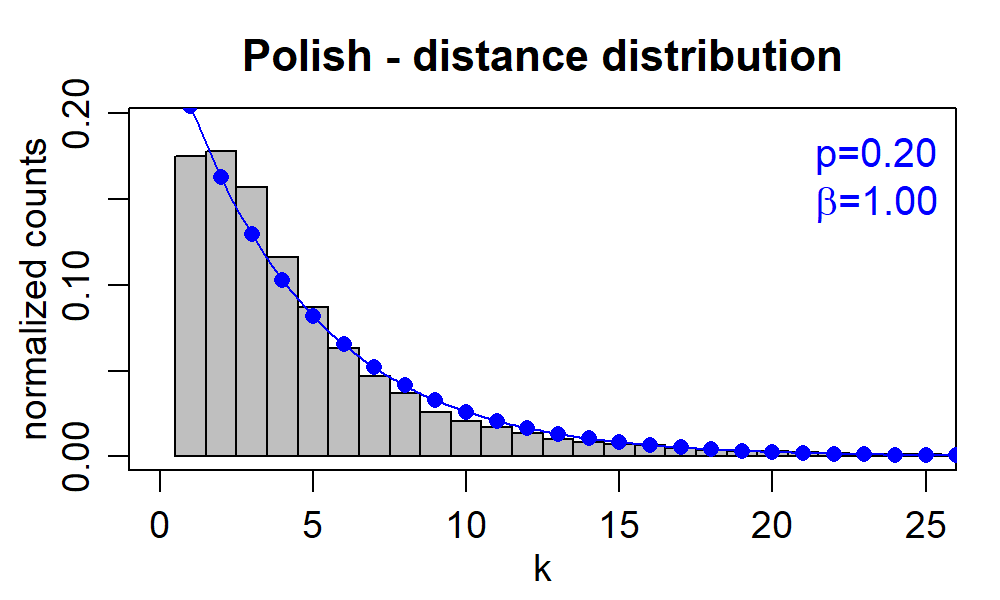}
\hspace{\figDistrSpace}
\includegraphics[width=\figDistrWidth]{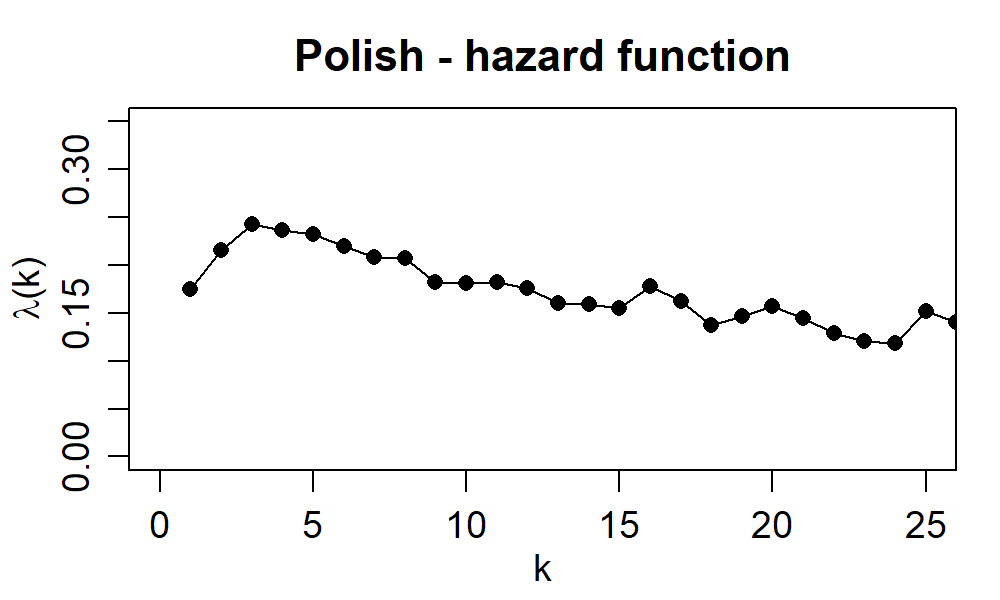}

\includegraphics[width=\figDistrWidth]{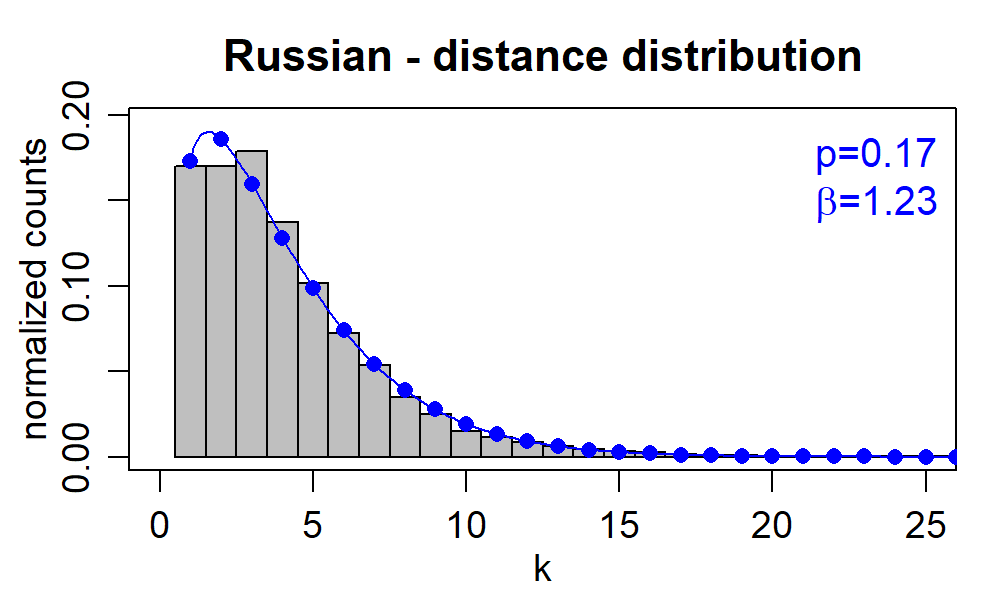}
\hspace{\figDistrSpace}
\includegraphics[width=\figDistrWidth]{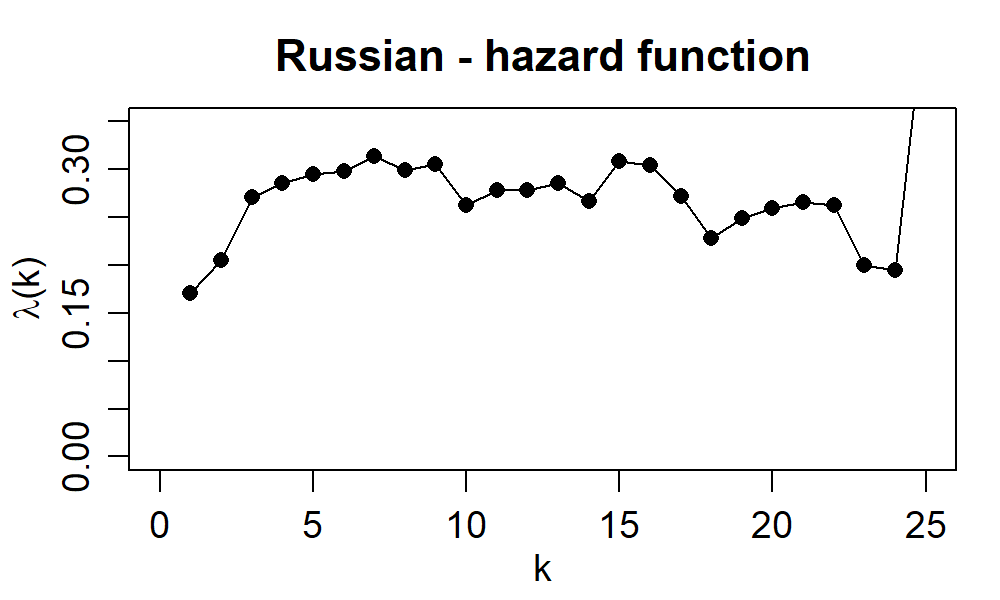}

\caption{The distributions of the distances between consecutive punctuation marks (left column) and the corresponding hazard functions (right column) in \textit{Finnegans Wake} and its translations.}
\label{fig::cpm.weibull.hazard}
\end{figure}

The results illustrating the characteristics of the distributions of distances between all consecutive punctuation marks for \textit{Finnegans Wake} in the original and its five above-mentioned translations are shown in Fig.~\ref{fig::cpm.weibull.hazard}. The panels in the left column show the corresponding normalized counts for increasing values of the distance $k$ measured by the number of words between consecutive punctuation marks. These empirical distributions are then fitted by the formula of Eq.~(\ref{eq::discrete.weibull.pmf}) and the parameters $p$ and $\beta$ of the best fits are explicitly given in the corresponding panels. For \textit{Finnegans Wake} in the original, it is a good fit, which was already presented in ref.~\cite{StaniszT-2024b}, but, remarkably, the parameter $\beta$ here is less than 1 in contrast to the multiplicity of all other literary texts studied~\cite{StaniszT-2023a} written in the major European languages, even those classified as belonging to experimental literature~\cite{StaniszT-2024b}. In general, the Weibull functional form describes well the distributions of distances between consecutive punctuation marks, but the corresponding parameters are somewhat different and on average are language-specific, so that they even undergo appropriate transformations in translations. The special case of \textit{Finnegans Wake} analyzed here, in its uniqueness, turns out to be much less susceptible to changing the values of these parameters. For translations into Dutch and French they even remain almost exactly unchanged. For the translation into Polish $\beta$ approaches the limit of 1 but does not exceed it, while for typical texts in this language $\langle \beta \rangle \approx 1.4$. For the two remaining translations, German and Russian, $\beta$ exceeds the value of 1, especially clearly for the Russian translation considered here, although in both of these cases it remains below the average value characteristic for these languages~\cite{StaniszT-2023a}. In this context, it should be noted here at the same time that the quality of the fit is slightly worse in these two cases, especially for small values of $k$.

A complementary insight into the specificity of the punctuation distribution is obtained through the hazard function $\lambda(k)$ defined by Eq.~(\ref{eq::hazard_function}). It gives a better perspective on the asymptotics of the distribution at larger values of $k$. These functions calculated directly from empirical data (not from the $p$ and $\beta$ fit parameters using the Weibull distribution) are presented in parallel on the right side of Fig.~\ref{fig::cpm.weibull.hazard}. As it can be seen from this perspective, all translations have a similar, decreasing asymptotics, with the Dutch and French translations having almost the same course as the English original. Apart from an initial slightly larger increase, the Polish translation behaves similarly. Such initial increases in $\lambda(k)$ are more pronounced for German and Russian translations, but for larger $k$ the decreases start to prevail even in these cases.

\subsection{Long-range correlations in IPI}

The distributions of fluctuations are one of the important characteristics of time series. With the same distributions, however, such series may differ in the arrangement of successive values in the series relative to each other. The MFDFA formalism presented above is well suited to quantifying possible long-range correlations of this type and is therefore applied to the time series of successive IPI's analyzed here, proceeding through the entire text of the book. 

\begin{figure}[p]
\newlength{\figTsAllpunctWidth}
\setlength{\figTsAllpunctWidth}{0.8\textwidth}
\centering

\includegraphics[width=\figTsAllpunctWidth]{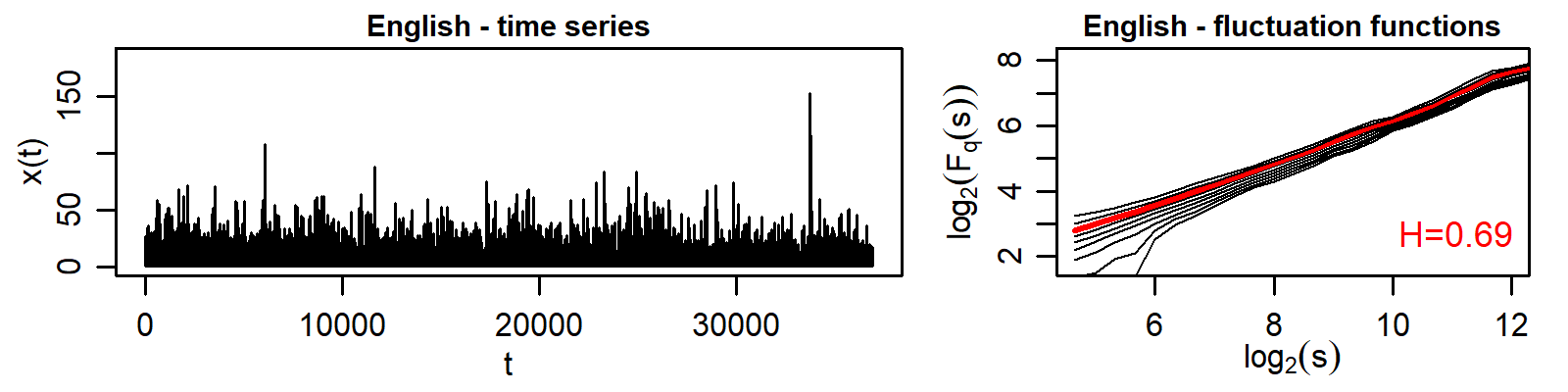}

\includegraphics[width=\figTsAllpunctWidth]{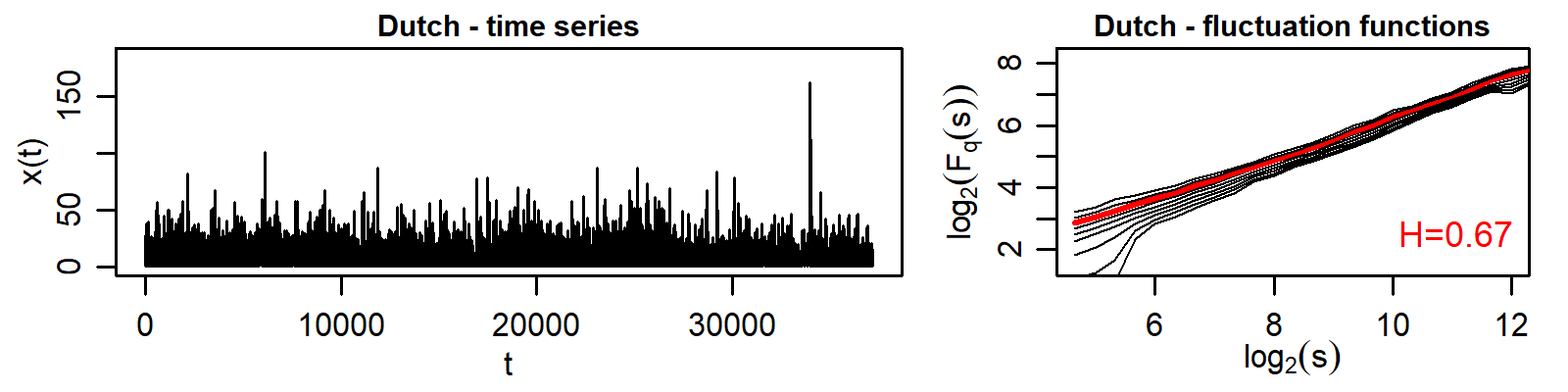}

\includegraphics[width=\figTsAllpunctWidth]{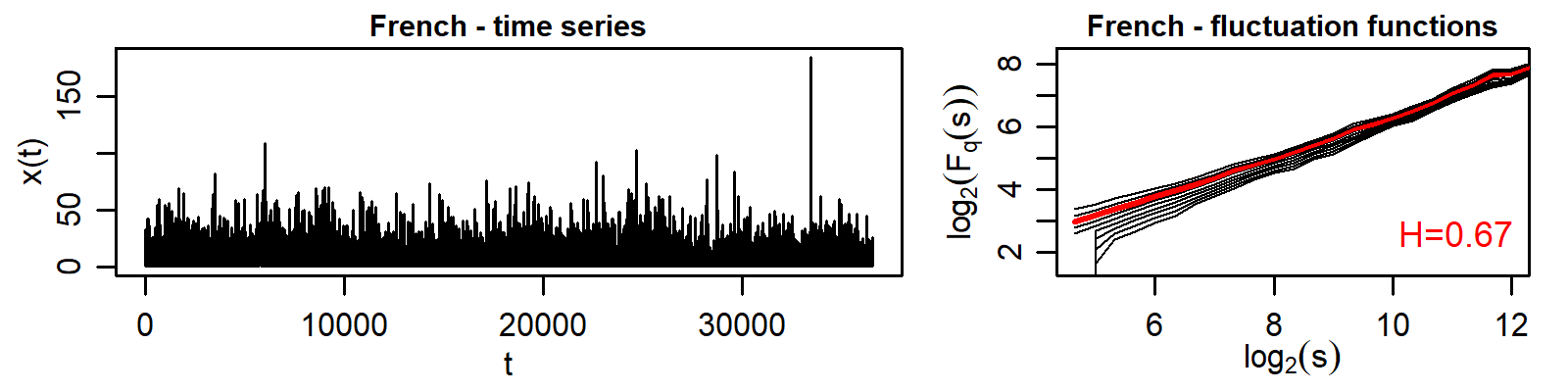}

\includegraphics[width=\figTsAllpunctWidth]{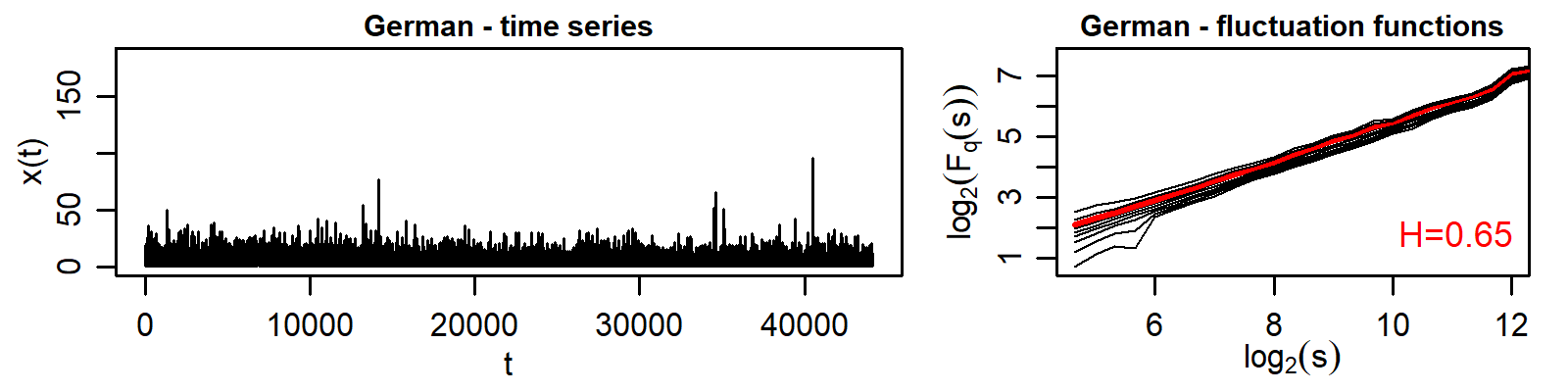}

\includegraphics[width=\figTsAllpunctWidth]{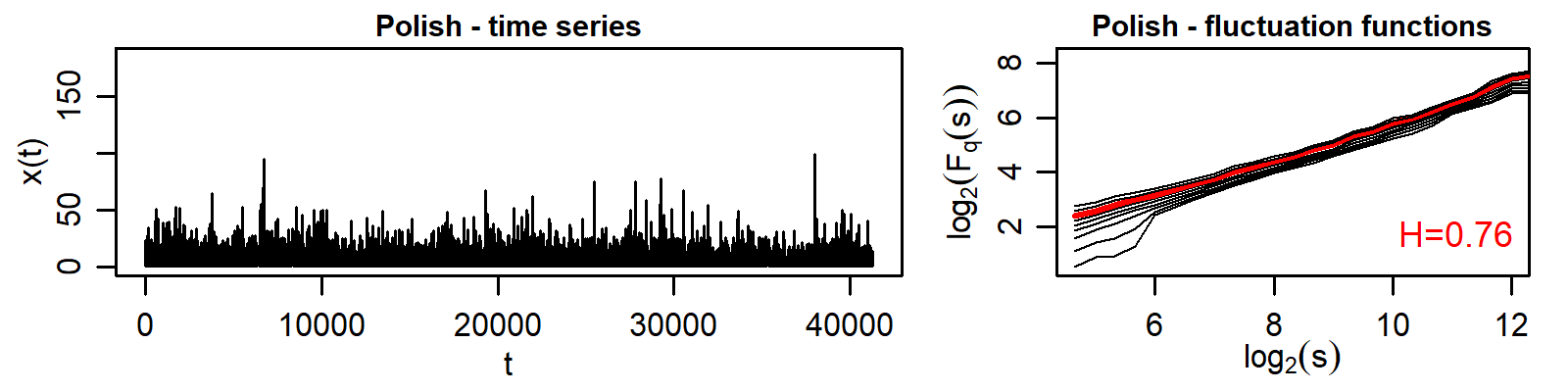}

\includegraphics[width=\figTsAllpunctWidth]{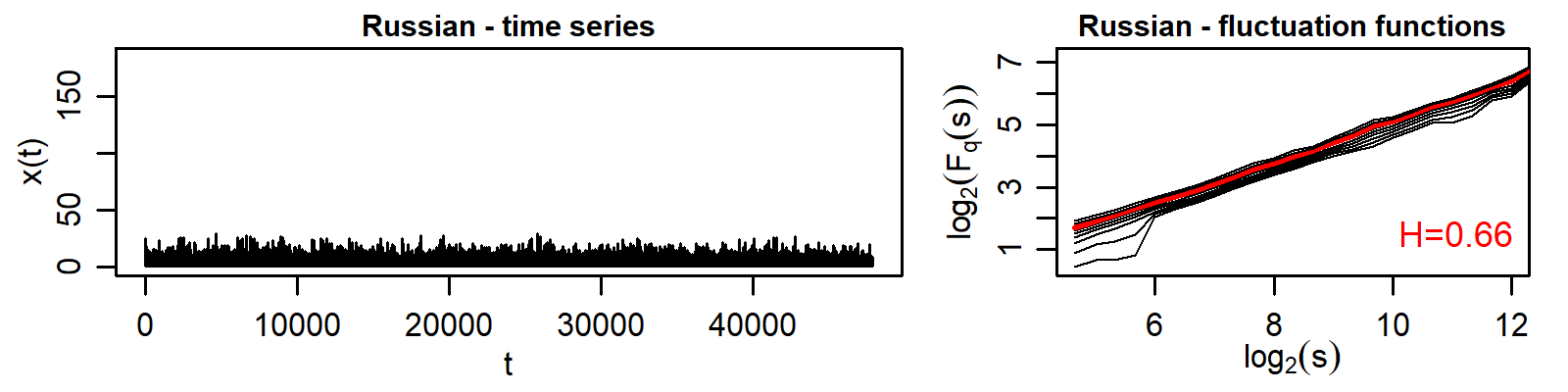}

\caption{Time series representing the distances between consecutive punctuation marks (left column) and the corresponding fluctuation functions (right column) in \textit{Finnegans Wake} and its translations. In each of the fluctuation functions plot, the function for $q=2$ is marked in red and the Hurst exponent $H$ is given in the bottom-right corner.}
\label{fig::cpm.mfdfa.hurst}
\end{figure}

The series of distances measured in word counts between all consecutive punctuation marks for the original \textit{Finnegans Wake} and its five translations are visualized in the left panels of Fig.~\ref{fig::cpm.mfdfa.hurst}. The same scale is used on the vertical axis to make the relative values directly visible. As it can be seen immediately, these series for the Dutch and French translations have strongly similar values and courses to those for the English original. The Polish translation is also quite consistent with these values. For the German translation some correspondence is also noticeable, but there are clearly more small numbers here and so the series is longer than in the above cases. An extreme case is the Russian translation, where the values are on average even significantly smaller, so their number in the series is even greater than for the German translation. 

The nature of the long-range correlations encoded in these series is quantified within MFDFA in terms of the fluctuations functions $F_q(s)$ expressed by the formula Eq.~(\ref{eq::fluctuation.functions}). These functions, calculated for $-4 \leq q \leq 4$ and plotted in the right-hand panels of Fig.~\ref{fig::cpm.mfdfa.hurst}, respectively, show clear scaling (straight line on the log-log scale) but this is not a convincing multifractal scaling because the dependence of these functions on $q$ is weak. These dependencies are rather close to monofractal. The slope of $F_q(s)$ lines is, however, clearly larger than $0.5$ and such are in particular the Hurst exponents $H=h(q=2)$, red line in this Figure) determined by Eq.~(\ref{eq::scaling}) and they are found to lie in the range $0.65 \leq H \leq 0.75$. This indicates clear long-range correlations of a persistent nature. The Polish translation, which is at the upper end of the $H$-value range, is the most extreme in this respect.

\begin{figure}[h]
\centering
\includegraphics[width=0.9\textwidth]{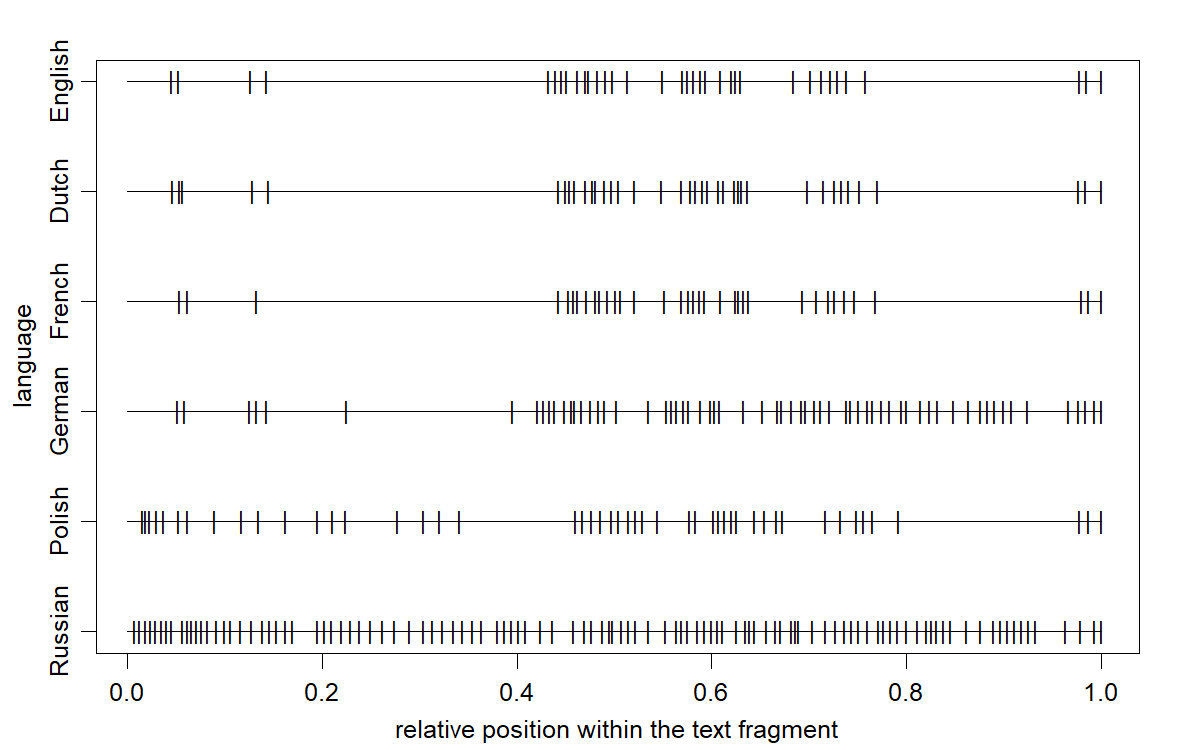}
\caption{The relative arrangement of punctuation marks corresponding to the six considered cases of the original \textit{Finnegans Wake} and its Dutch, French, German, Polish, and Russian translations, in a two-paragraph excerpt from the book (starting at the beginning of the last paragraph on page 579 and ending at the end of page 580 in the original, English version).}
\label{fig::punctuation.excerpt}
\end{figure}

The illustration that complements this section, shown in Fig.~\ref{fig::punctuation.excerpt}, explicitly demonstrates one of the extreme examples of differences in the distribution of punctuation in some translations of \textit{Finnegans Wake}. The plot presents the relative arrangement of punctuation marks in a sample two-paragraph excerpt from the book. The discussed excerpt starts at the beginning of the last paragraph on page 579 and ends at the end of page 580 in the original, English version. Clearly, consistent with the previous observations, the first three cases (English, Dutch, and French) differ least while the Russian translation differs most by generating many much shorter IPI's. 

\subsection{Multifractal sentence length variability (SLV)}

The distribution of distances between all punctuation marks clearly follows certain fairly universal rigors, which have already been shown by previous recent studies and which are confirmed to a large extent even for such an exceptional work as the one examined here, \textit{Finnegans Wake}. At the same time, however, the distribution of the punctuation marks ending sentences (such as periods, question marks, and exclamation marks) have much more freedom in this respect. The distances between them are precisely the lengths of sentences. It has already been shown~\cite{DrozdzS-2016a} that \textit{Finnegans Wake} develops a self-similar cascade of SLV with exceptionally richly developed multifractal properties. Such patterns appear only in literary works written using the narrative technique of stream of consciousness, and \textit{Finnegans Wake} is the most spectacular in this respect, both in terms of an almost ideal symmetry of the resulting singularity spectrum (Eq.~(\ref{eq::singularity.spectrum})) as well as in the sense of the width of this spectrum. The main goal of the present study is to examine to what extent these properties are preserved in the translations available here.

The results collected in multi-panel Fig.~\ref{fig::cpm.mfdfa.multifractal.1} clearly indicate that, in this respect, the translations largely faithfully preserve the multifractal characteristics of the original, and in the two cases of French and Polish this mapping is almost perfect. The patterns of variation of successive sentence lengths are strikingly similar here, and this applies even to their absolute values, as can be seen on the vertical scales of the upper panels of this figure. Consequently, the total number of sentences is also similar, as it can be seen on the horizontal scales of these panels. This applies even to the Russian translation, which showed greater discrepancies in the case of full punctuation. The functions $F_q(s)$ show very good scaling strongly dependent on $q$, comparably in all cases. As a consequence, the multifractal spectra $f(\alpha)$ are alike, broad ($\Delta \alpha \approx 1$), and vary only slightly. Minor asymmetries in the spectra can also be observed. For Dutch and Russian, they are right-sided and for German they are left-sided. This indicates~\cite{DrozdzS-2015a} that, in the first two cases, the hierarchy of multifractal correlations is somewhat more developed towards variability of shorter sentence lengths than longer ones, while in the third case (German) it is the other way around. The multifractal spectra of the French and Polish translations are almost exactly symmetrical, similarly to the English original. All these spectra are also significantly shifted to the right with respect to $\alpha = 0.5$, which signals a clear persistent trend in sentence length variability. This property is also confirmed in terms of the corresponding Hurst exponents $H$, which are calculated as scaling exponents for $F_{q=2}(s)$, marked in red in these Figures. Their corresponding values turn out to be even slightly higher than those when considering the variation of distances between all successive punctuation marks, which has been marked in Fig.~\ref{fig::cpm.mfdfa.hurst}.

\begin{figure}[p]
\newlength{\figMfdfaSentWidth}
\setlength{\figMfdfaSentWidth}{0.8\textwidth}
\centering

\includegraphics[width=\figMfdfaSentWidth]{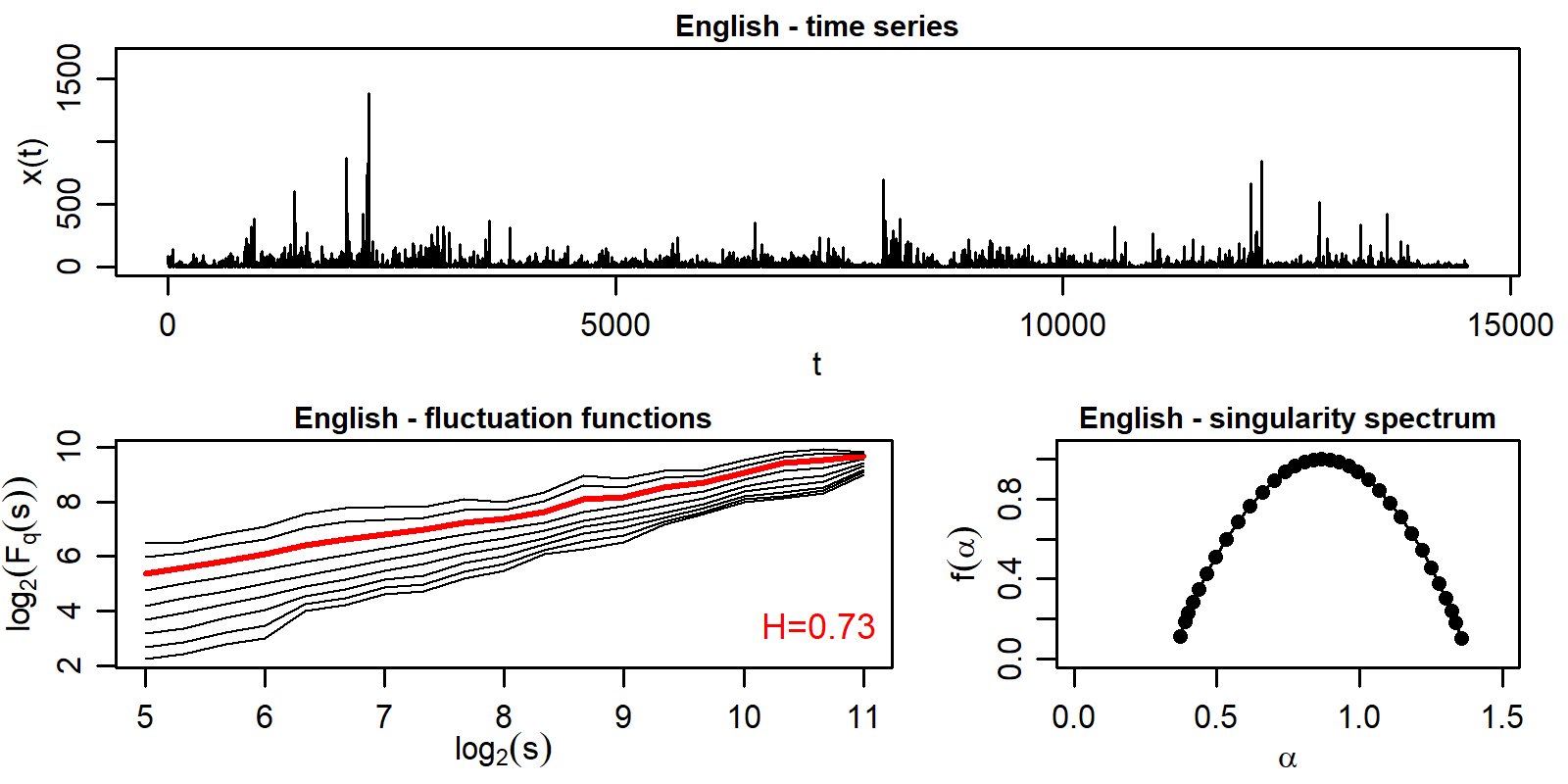}

\includegraphics[width=\figMfdfaSentWidth]{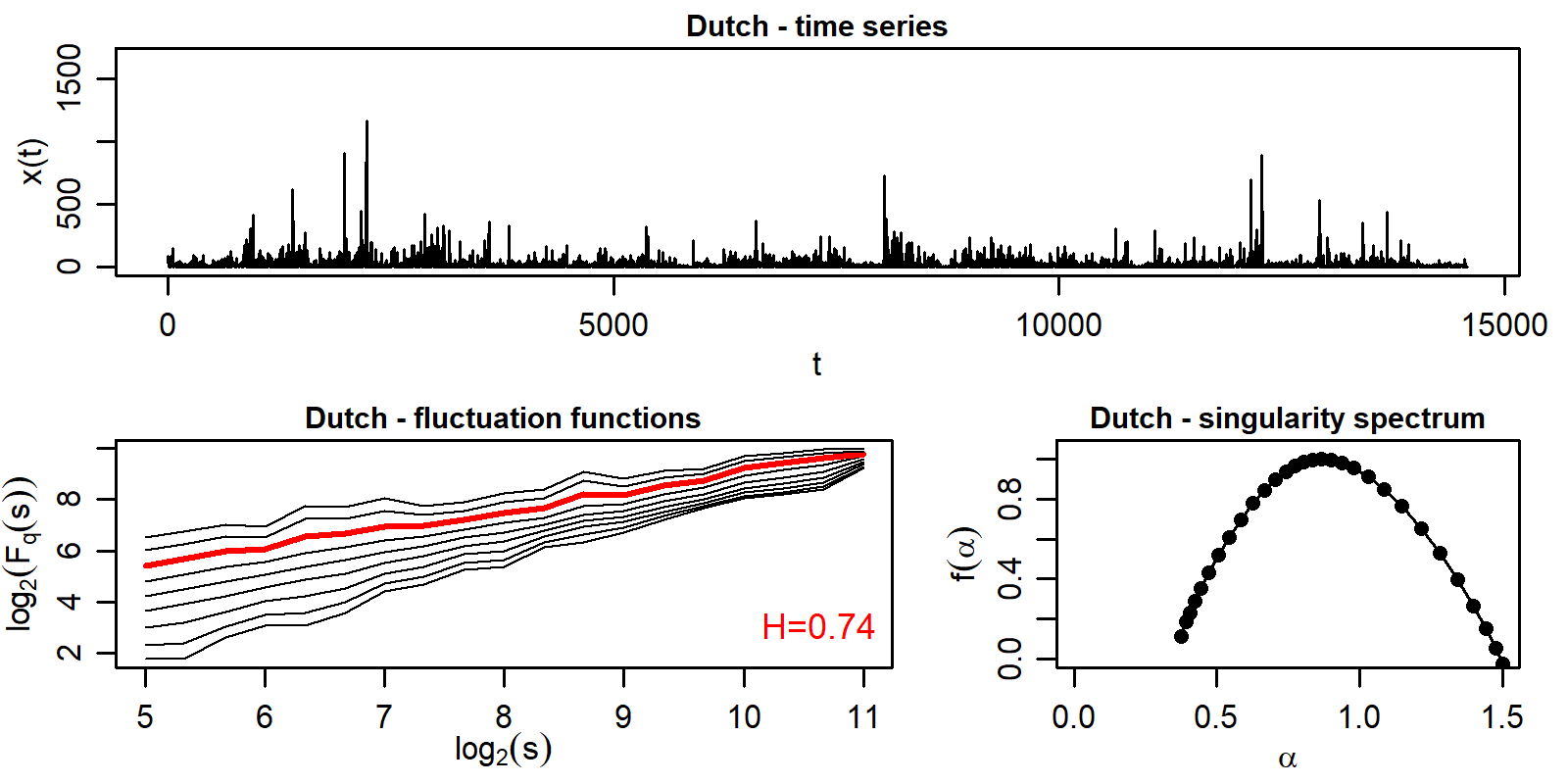}

\includegraphics[width=\figMfdfaSentWidth]{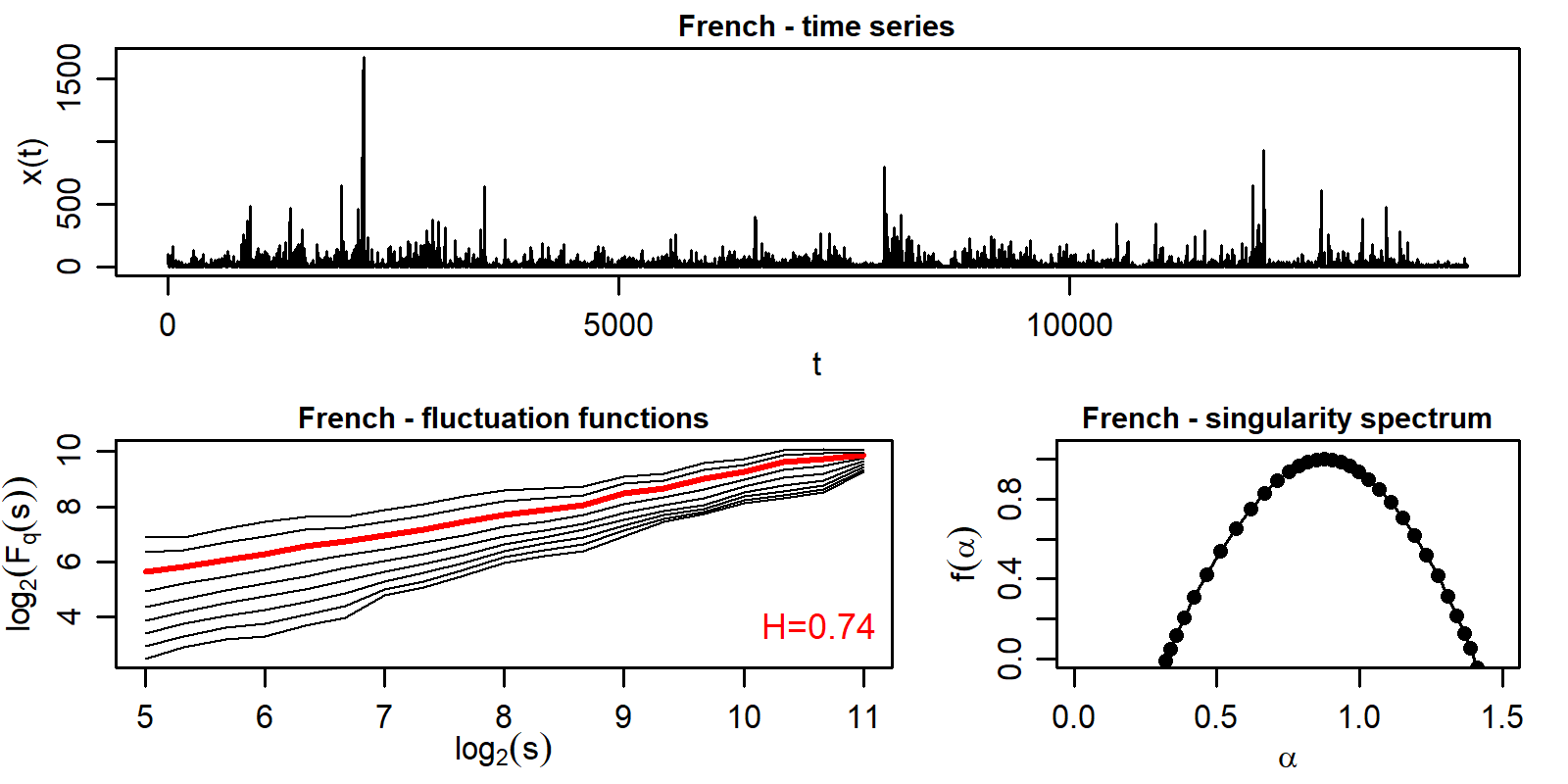}

\caption{Time series representing sentence lengths, the corresponding fluctuation functions $F_q(s)$, and singularity spectra $f(\alpha)$ in \textit{Finnegans Wake} and its translations. In each of the fluctuation function plots, $F_q(s)$ for $q=2$ is marked in red and the Hurst exponent $H$ value is given in the bottom-right corner.}
\label{fig::cpm.mfdfa.multifractal.1}
\end{figure}

\begin{figure}[p]
\ContinuedFloat
\setlength{\figMfdfaSentWidth}{0.8\textwidth}
\centering

\includegraphics[width=\figMfdfaSentWidth]{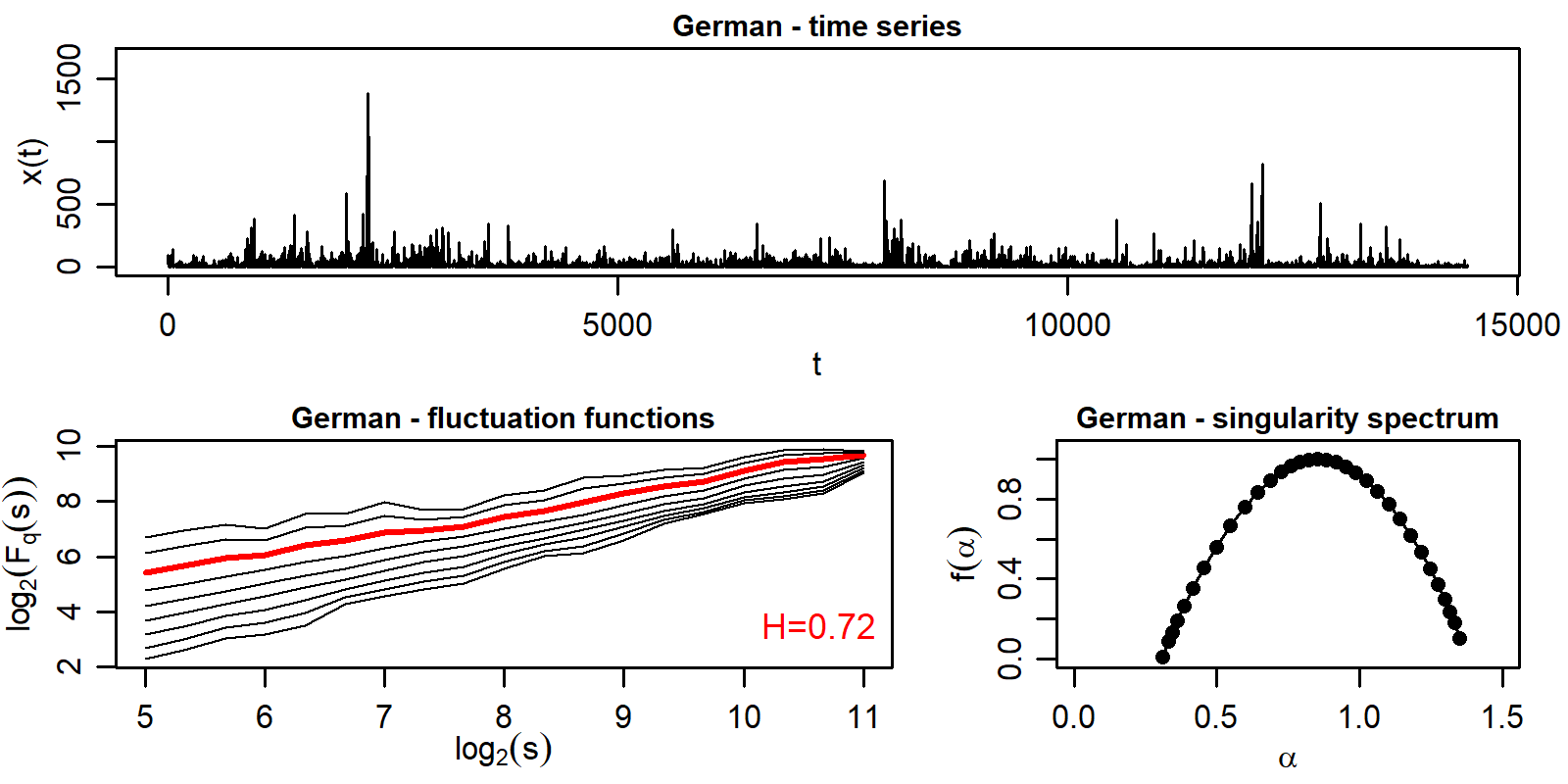}

\includegraphics[width=\figMfdfaSentWidth]{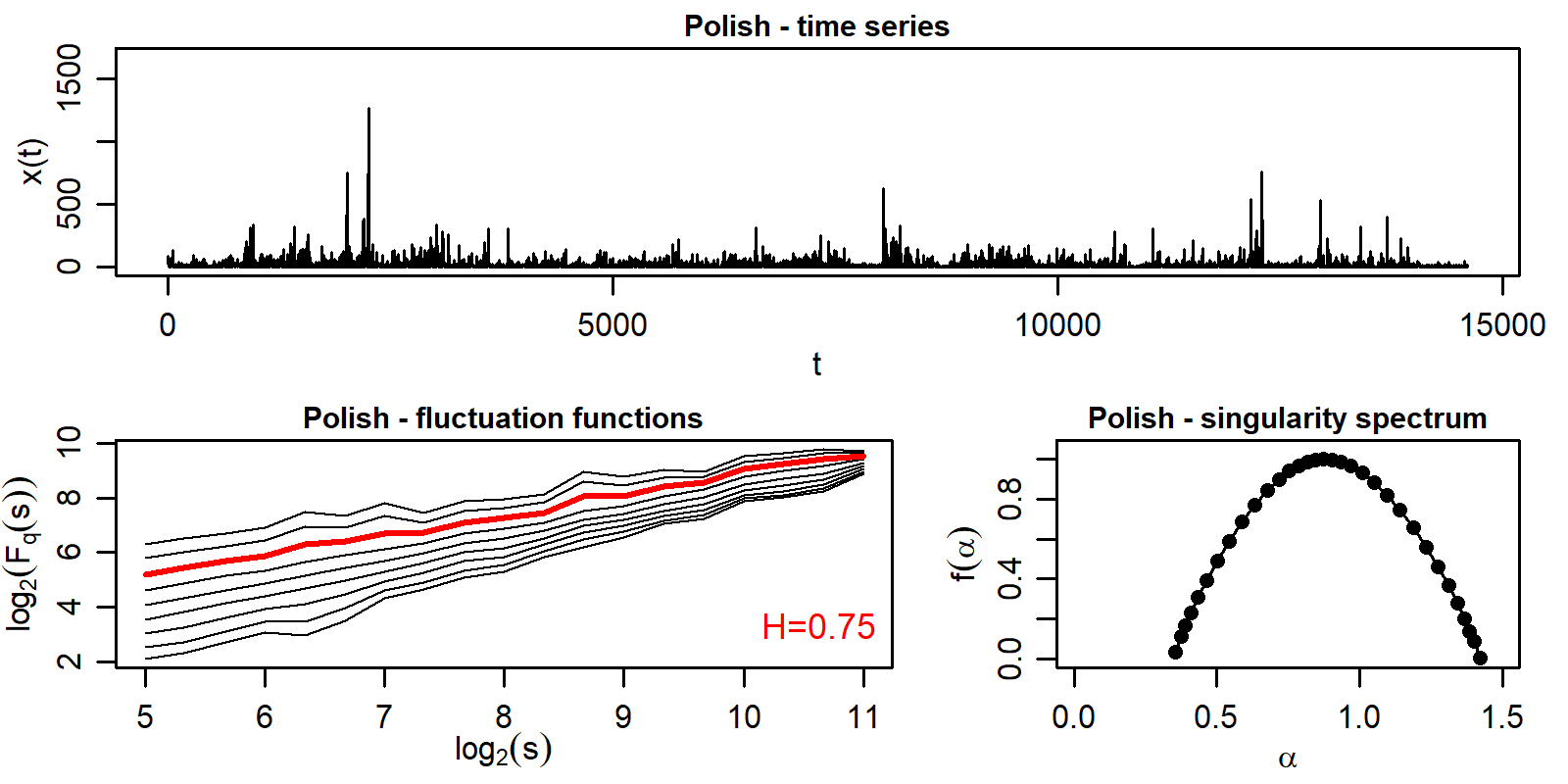}

\includegraphics[width=\figMfdfaSentWidth]{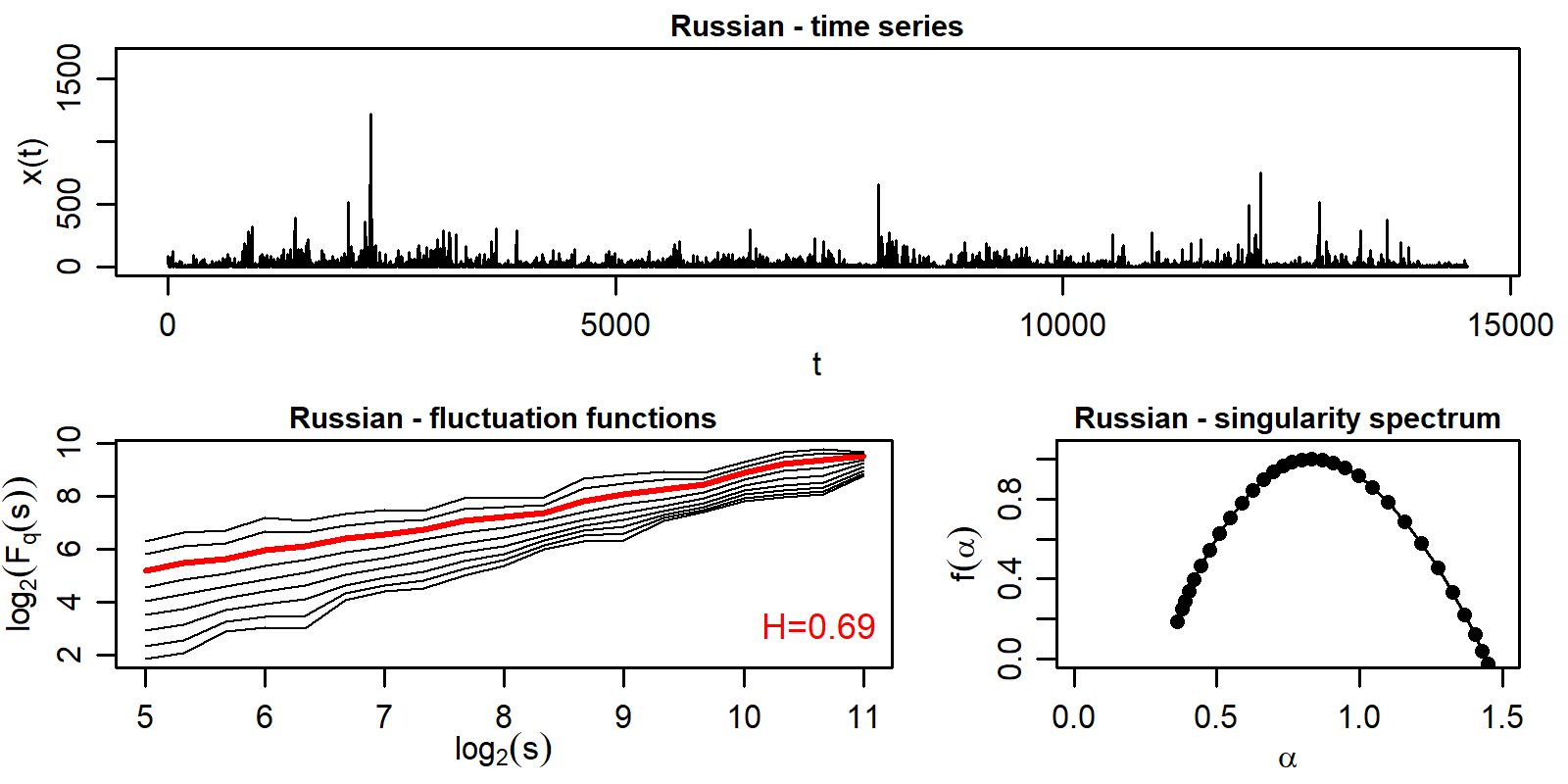}

\caption{(continued) Time series representing sentence lengths, the corresponding fluctuation functions $F_q(s)$, and singularity spectra $f(\alpha)$ in \textit{Finnegans Wake} and its translations. In each of the fluctuation function plots, $F_q(s)$ for $q=2$ is marked in red and the Hurst exponent $H$ value is given in the bottom-right corner.}
\label{fig::cpm.mfdfa.multifractal.2}
\end{figure}

\section{Summary and conclusions}

This study explored the intricate punctuation patterns and sentence length variability within James Joyce's \textit{Finnegans Wake} and its translations into five languages: Dutch, French, German, Polish, and Russian. Through a combination of statistical methods, including the discrete Weibull distribution and multifractal detrended fluctuation analysis (MFDFA), it is identified that punctuation usage in \textit{Finnegans Wake} defies conventional linguistic norms. Unlike typical texts, its punctuation patterns remain largely translation-invariant, reflecting Joyce's intentional crafting of a translinguistic narrative.

The findings highlight two distinct characteristics. The text exhibits a rare decreasing hazard function in its punctuation intervals, a trait only observed in \textit{Finnegans Wake} among analyzed works. This uniqueness persists across most translations, with minor variations reflecting linguistic differences, or, what cannot be excluded, some bias related to the translation fidelity. The multifractal properties of sentence lengths in \textit{Finnegans Wake} demonstrate an extraordinary degree of self-similarity and complexity. This trait is preserved remarkably well in translations, particularly in French and Polish versions, underscoring the robustness of the text’s structural organization. These results contribute to a deeper understanding of the interplay between linguistic structure and translation fidelity, affirming the universal complexity embedded in Joyce's work.

The study confirms that \textit{Finnegans Wake} exemplifies a rare literary phenomenon where structural properties transcend linguistic boundaries, maintaining coherence in complexity across translations. This underscores the text’s suitability for cross-disciplinary analysis, superimposing linguistics, literature, and complexity science. By demonstrating that punctuation and sentence organization are integral to the text’s identity, this research highlights the intricate balance between translation and preservation of artistic intent. 

Future research could extend these methods to other experimental literary works to further explore the universality of such translinguistic traits. Additionally, integrating advanced computational techniques and larger datasets might provide new insights into the quantitative characteristics of experimental literature, which explores the deeper layers and possibilities of natural language.

\authorcontributions{Conceptualization, K.B., S.D., J.K. and T.S.; Methodology, S.D., J.K. and T.S.; Software, J.K. and T.S.; Validation, K.B., S.D., J.K. and T.S.; Formal analysis, S.D., J.K. and T.S.; Investigation, S.D., J.K. and T.S.; Resources, K.B. and T.S.; Data curation, K.B. and T.S.; Writing---original draft, S.D.; Writing---review \& editing, K.B., S.D., J.K. and T.S.; Visualization, J.K. and T.S.; Supervision, S.D.}

\funding{This research received no external funding.}

\institutionalreview{Not applicable.}

\dataavailability{The data presented in this study are available on request from the corresponding author. The data are not publicly available due to copyright.}

\conflictsofinterest{The authors declare no conflicts of interest.}

\begin{adjustwidth}{-\extralength}{0cm}
\reftitle{References}

\bibliography{bibliography_bibtex_file}

\PublishersNote{}
\end{adjustwidth}
\end{document}